\documentclass[11pt,a4paper]{article}
\pdfoutput=1
\usepackage[table]{xcolor}
\usepackage[hyperref]{naaclhlt2018}
\usepackage{times}
\usepackage{latexsym}
\usepackage{url}
\usepackage{amsmath}
\usepackage{amsfonts}
\usepackage{amssymb}
\usepackage[safe]{tipa}
\usepackage{xspace}
\usepackage{microtype}
\usepackage{xfrac}
\usepackage{nicefrac}
\usepackage{booktabs}
\usepackage{color}
\usepackage{adjustbox}
\usepackage{graphicx}
\usepackage[small]{caption}
\usepackage{subcaption}

\definecolor{darkred}{rgb}{0.55, 0.0, 0.0}
\definecolor{darkblue}{rgb}{0.0, 0.0, 0.55}
\definecolor{darkmagenta}{rgb}{0.55, 0.0, 0.55}
\definecolor{darkgreen}{rgb}{0.01, 0.75, 0.24}
\definecolor{darkyellow}{rgb}{1.0, 0.87, 0.0}

\newcommand{\oov}{\textsc{unk}\xspace}
\newcommand{\eos}{\textsc{eos}\xspace}
\newcommand{\eow}{\underline{\textsc{eow}}\xspace}

\newcommand{\word}[1]{\textit{#1}}
\newcommand{\lex}[1]{\textsc{#1}}
\newcommand{\LM}{LM\xspace}
\newcommand{\LMs}{LMs\xspace}
\newcommand{\question}{\scalebox{1.2}{$\star$}}
\newcommand{\charseq}{{\boldsymbol c}}

\newcommand{\cupdot}{\mathbin{\raisebox{-.25em}{$\dot{\smash{\raisebox{.25em}{$\cup$}}}$}}}

\usepackage{cleveref}
\crefname{section}{\S}{\S\S}
\Crefname{section}{\S}{\S\S}
\crefname{table}{Table}{}
\crefname{figure}{Fig.}{}
\crefname{algorithm}{alg.}{}
\crefname{equation}{eq.}{}
\crefname{appendix}{App.}{}
\crefformat{section}{\S#2#1#3}  

\aclfinalcopy 
\def\aclpaperid{995} 

\title{Are All Languages Equally Hard to Language-Model?}

\author{Ryan Cotterell$^1$ {\;\normalfont and\;} Sabrina J. Mielke$^1$ {\;\normalfont and\;} Jason Eisner$^1$ {\;\normalfont and\;} Brian Roark$^2$ \\
  ${}^1$ Department of Computer Science, Johns Hopkins University  \hspace*{0.3in}
  ${}^2$ Google \\
  {\tt \{ryan.cotterell@,sjmielke@,jason@cs.\}jhu.edu} \;\; {\tt roark@google.com} \\}

\date{}

\begin{document}

\thispagestyle{plain}
\pagestyle{plain}

\maketitle
\begin{abstract}
  For general modeling methods applied to diverse languages, a natural
question is: how well should we expect our models to work on languages
with differing typological profiles?   In this work, we develop an
  evaluation framework for fair cross-linguistic comparison of
  language models, using translated text so that all models are
  asked to predict approximately the same information.
  We then conduct a study on 21 languages, demonstrating that
  in some languages, the textual expression of the information
  is harder to predict with both $n$-gram and LSTM language
  models.
  We show complex inflectional morphology
  to be a cause of performance differences among
  languages.
\end{abstract}

\section{Introduction}

Modern natural language processing practitioners strive to create
modeling techniques that work well on all of the world's
languages. Indeed, most methods are portable in the following sense:
Given appropriately annotated data, they should, in principle, be
trainable on any language. However, despite this crude
cross-linguistic compatibility, it is unlikely that all languages
are equally easy, or that our methods are equally good at all languages.

In this work, we probe the issue, focusing on
{\em language modeling}.  A fair comparison is tricky.
Training corpora in different languages have different sizes, and
reflect the disparate topics of discussion in different linguistic communities, some of which may be
harder to predict than others. Moreover, bits per character, a
standard metric for language modeling, depends on the vagaries of a
given orthographic system. We argue for a fairer metric based on the
bits per utterance using utterance-aligned multi-text.
That is, we train and test on ``the same'' set of utterances in each
language, modulo translation. To avoid discrepancies in
out-of-vocabulary handling, we evaluate open-vocabulary models.

We find that under standard approaches, text tends to be harder to
predict in languages with fine-grained inflectional morphology.
Specifically, language models perform worse on these languages,
in our controlled comparison.
Furthermore, this performance difference essentially vanishes when we
remove the inflectional markings.\footnote{One might have expected
  {\em a priori} that some difference would remain, because most highly
  inflected languages can also vary word order to mark a topic-focus
  distinction, and this (occasional) marking is preserved in our
  experiment.}

Thus, in highly inflected languages, either the utterances have
more content or the models are worse.
(1) Text in highly inflected languages may be {\em inherently harder
  to predict} (higher entropy per utterance) if its extra morphemes
carry additional, unpredictable information.  (2) Alternatively,
perhaps the extra morphemes are {\em predictable in principle}---for
example, redundant marking of grammatical number on both subjects and
verbs, or marking of object case even when it is predictable from
semantics or word order---and yet our current language modeling
technology fails to predict them.  This might happen because (2a) the
technology is biased toward modeling words or characters and fails to
discover intermediate morphemes, or because (2b) it fails to capture
the syntactic and semantic predictors that govern the appearance of
the extra morphemes.  We leave it to future work to tease apart these
hypotheses.

\section{Language Modeling}
A traditional closed-vocabulary, word-level language model operates as
follows: Given a fixed set of words ${\cal V}$, the model provides a
probability distribution over sequences of words with parameters to
be estimated from data.
Most fixed-vocabulary language models employ a distinguished symbol
\oov that represents all words not present in ${\cal V}$;
these words are termed out-of-vocabulary (OOV).

Choosing the set ${\cal V}$ is something of a black art: Some
practitioners choose the $k$ most common words
(e.g., \newcite{DBLP:conf/interspeech/MikolovKBCK10} choose $k=10000$) and
others use all those words that appear at least twice in the training
corpus. In general, replacing
more words with \oov artificially improves the perplexity measure
but produces a less useful model. OOVs
present something of a challenge for the cross-linguistic comparison
of language models, especially in morphologically rich languages, which simply
have more word forms.

\begin{figure*}
    \centering
    \begin{subfigure}[t]{0.31\textwidth}
        \includegraphics[width=\textwidth]{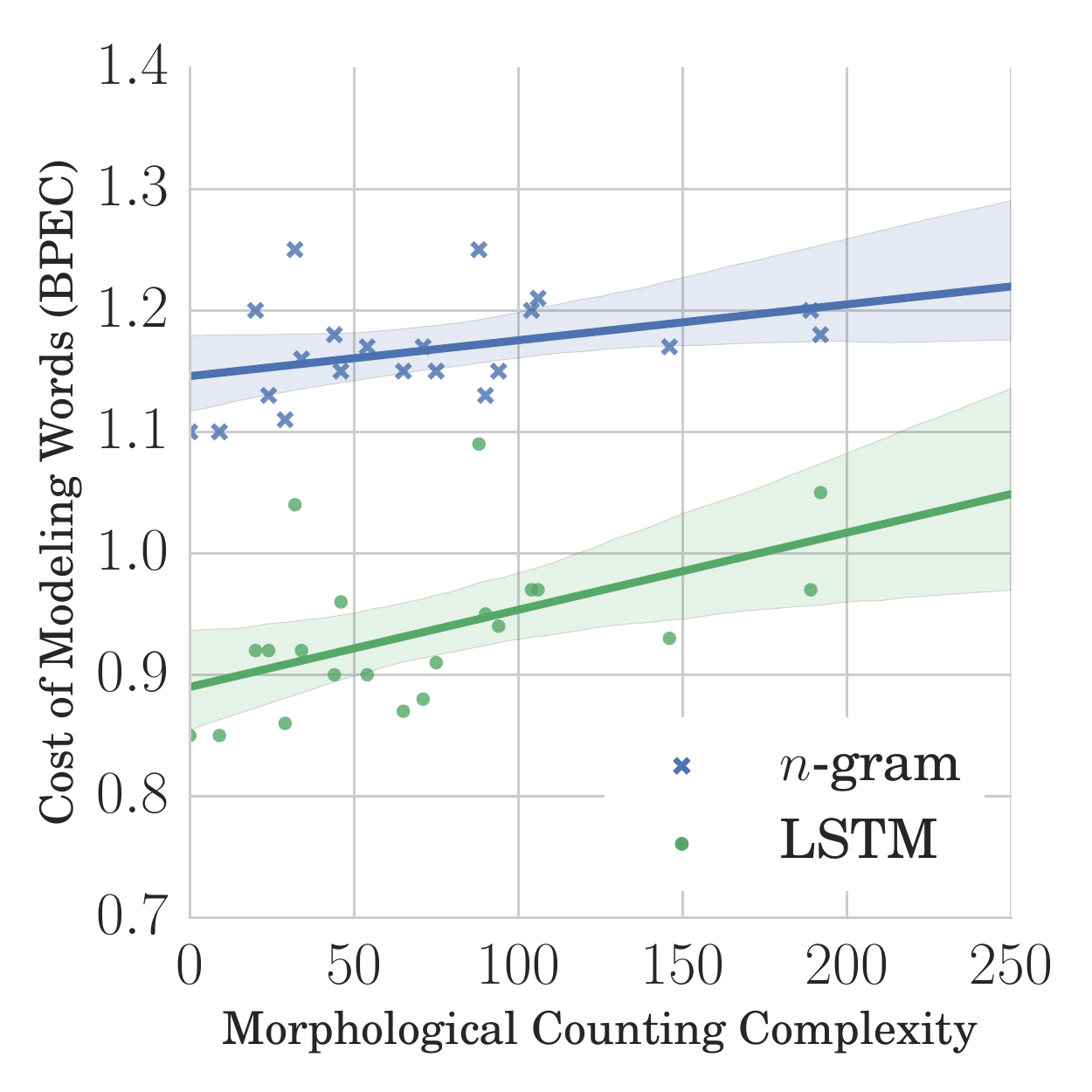}
        \caption{BPEC performance of $n$-gram (blue) and LSTM
         (green) {\LM}s over word sequences. Lower is better.}
        \label{fig:words}
    \end{subfigure}
    ~ 
    \begin{subfigure}[t]{0.31\textwidth}
        \includegraphics[width=\textwidth]{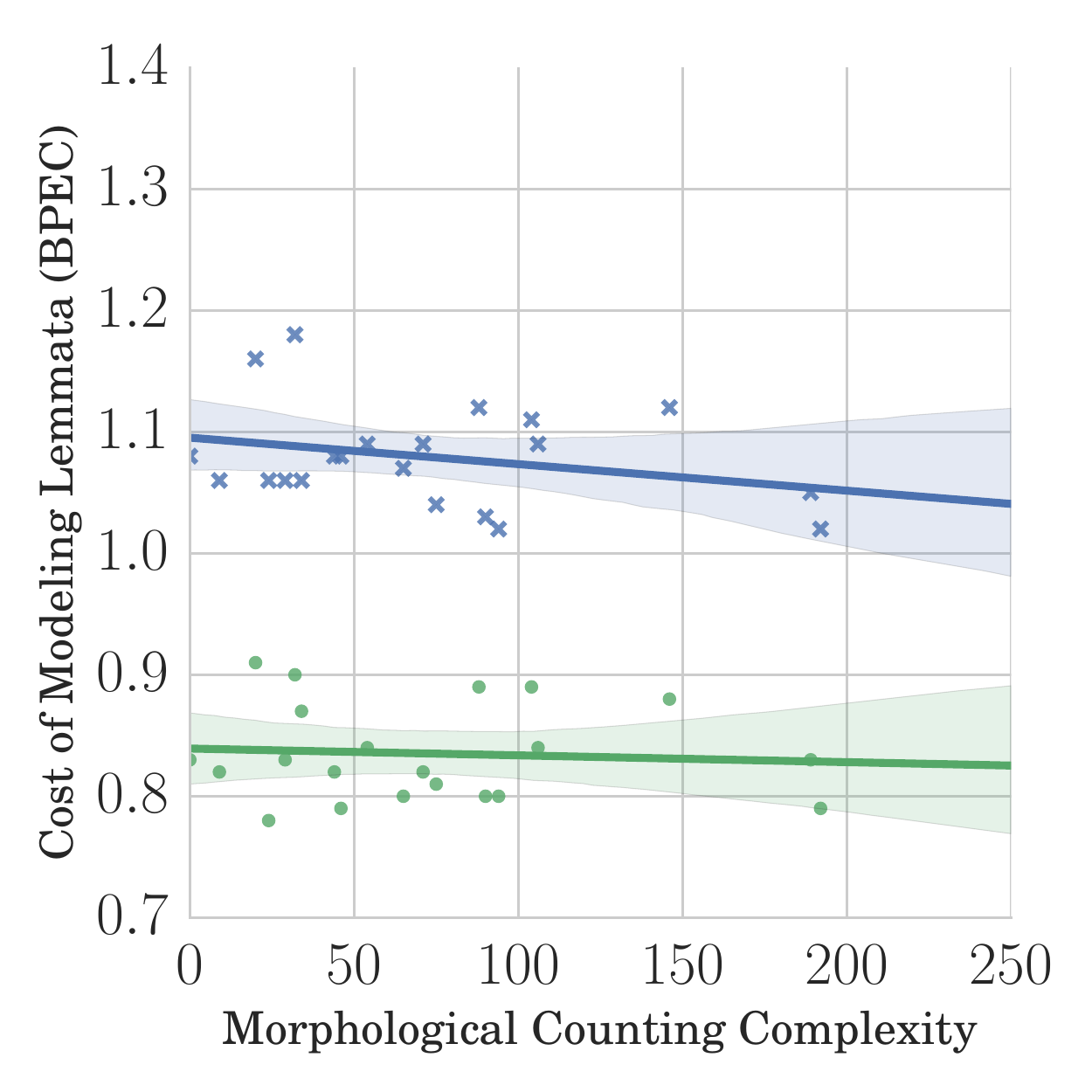}
        \caption{BPEC performance of $n$-gram (blue) and LSTM
        (green) {\LM}s over lemma sequences.  Lower is better.}
        \label{fig:lemmata}
    \end{subfigure}
    ~ 
    \begin{subfigure}[t]{0.31\textwidth}
        \includegraphics[width=\textwidth]{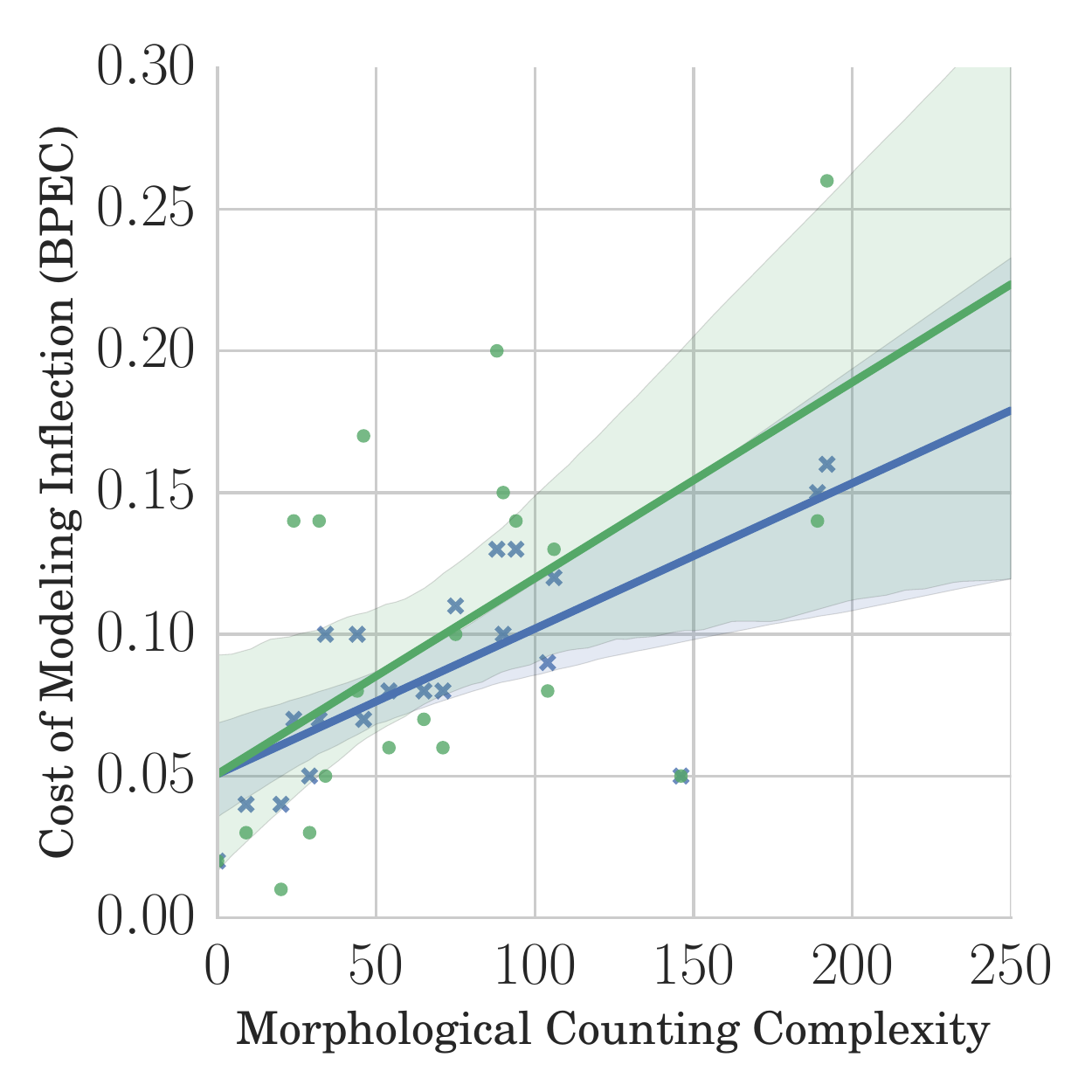}
        \caption{Difference in BPEC performance of $n$-gram (blue) and LSTM
          (green) {\LM}s between words and lemmata.}
        \label{fig:diff}
    \end{subfigure}
    \caption{The primary findings of our paper are evinced in these
      plots. Each point is a language. While the LSTM outperforms the
      hybrid $n$-gram model, the relative performance on the highly
      inflected languages compared to the more modestly inflected
      languages is almost constant; to see this point, note that the regression lines in
      \cref{fig:diff} are almost identical. Also, comparing \cref{fig:words} and \cref{fig:lemmata} shows that
      the correlation between \LM performance and morphological richness disappears after lemmatization of the corpus, indicating
    that inflectional morphology is the origin for the lower BPEC. }\label{fig:graphs}\vspace*{-0.12in}
\end{figure*}

\subsection{The Role of Inflectional Morphology}\label{sec:inflection}
Inflectional morphology can explode the base vocabulary of a
language. Compare, for instance, English and Turkish.  The nominal
inflectional system of English distinguishes two forms: a singular and
plural. The English lexeme
\lex{book} has the singular form \word{book} and the plural form
\word{books}. In contrast, Turkish distinguishes at least
12: \word{kitap}, \word{kitablar},
\word{kitab{\i}}, \word{kitab{\i}n}, etc.

To compare the degree of morphological inflection in our evalation
languages, we use \textbf{counting complexity}
\cite{sagot2013comparing}. This crude metric counts the number of
inflectional categories distinguished by a language (e.g., English includes
a category of 3rd-person singular present-tense verbs).  We
count the categories annotated in the language's UniMorph \cite{KIROV18} lexicon. See \cref{tab:bigv2} for the counting complexity of evaluated languages.

\subsection{Open-Vocabulary Language Models}

To ensure comparability across languages, we require our language models to predict every character in an utterance, rather than skipping some characters because they appear in words that were (arbitrarily) designated as OOV in that language.  Such models are known as ``open-vocabulary'' LMs.

\paragraph{Notation.}
Let $\cupdot$ denote disjoint union, i.e., $A \cupdot B = C$ iff $A \cup B = C$ and $A \cap B = \emptyset$.
Let $\Sigma$ be a discrete alphabet of characters, including a distinguished unknown-character symbol
$\question$.\footnote{The set of graphemes in these languages can be assumed to be closed,
but external graphemes may on rare occasion appear in random text
samples.  These are rare enough to not materially affect the metrics.}
A character \LM then defines
  $p(\charseq) = \prod^{|\charseq|+1}_{i=1} p(c_i \mid \charseq_{<i})$,
where we take $c_{|\charseq|+1}$ to be a distinguished end-of-string symbol \eos.
In this work, we consider two open-vocabulary LMs, as follows.

\paragraph{Baseline $n$-gram \LM.}
We train ``flat'' hybrid word/character open-vocabulary
$n$-gram models \cite{bisani2005open}, defined over strings $\Sigma^+$
from a vocabulary $\Sigma$ with mutually disjoint subsets:
$\Sigma = W \cupdot C \cupdot S$, where single characters $c\in C$ are distinguished in the
model from single character full words $w\in W$, e.g., \underline{\tt a}
versus the word {\tt a}.
Special symbols $S=\{\eow,\eos\}$ are end-of-word and end-of-string,
respectively. N-gram histories in $H$ are either word-boundary or
word-internal (corresponding to a whitespace tokenization), i.e., $H = H_b \cupdot H_i$.
String-internal word boundaries are always separated by a single whitespace character.\footnote{The model can be extended to
  handle consecutive whitespace characters or punctuation at word
  boundaries; for this paper, the tokenization split punctuation from
  words and reduced consecutive whitespaces to one, hence the simpler
  model.} For example, if $\mbox{{\tt foo}, {\tt baz}}\in W$ but
$\mbox{{\tt bar}}\not\in W$, then the string {\tt foo bar baz} 
would be generated as: \texttt{foo \underline{b} \underline{a} \underline{r}}\;\;\eow \, \texttt{baz} \, \eos.
Possible 3-gram histories in this string would be, e.g., [\texttt{foo \underline{b}}] $\in H_i$, [\texttt{\underline{r}}\;\;\eow\!] $\in H_b$, and [\eow \, \texttt{baz}] $\in H_b$.

Symbols are generated from a multinomial given the history $h$,
leading to a new history $h'$ that now includes the symbol and is
truncated to the Markov order. Histories $h\in H_b$ can generate
symbols $s\in W\cup C \cup \{\eos\}$. If $s = \eos$,
the string is ended. If $s \in W$, it has an implicit
\eow and the model transitions to history $h'\in H_b$. If $s\in C$, it
translitions to $h' \in H_i$.  Histories $h \in H_i$ can generate
symbols $s\in C \cup \{\eow\}$ and transition to
$h' \in H_b$ if $s = \eow$, otherwise to $h'\in H_i$.

We use standard \newcite{DBLP:conf/icassp/KneserN95} model training,
with distributions at word-internal histories $h \in H_i$ constrained
so as to only provide probability mass for symbols
$s\in C \cup \{\eow\}$. We train 7-gram models, but prune $n$-grams $h$$s$ where the
history $h\in W^k$, for $k>4$, i.e., 6- and 7-gram histories must include at least one
$s\not\in W$.
To establish the vocabularies $W$ and $C$, we
replace exactly one instance of each word type with its spelled out
version. Singleton words are thus excluded from $W$, and character
sequence observations from all types are included in training. Note
any word $w\in W$ can also be generated as a character sequence. For
perplexity calculation, we sum the probabilities for each way of
generating the word.

\paragraph{LSTM \LM.}
While neural language models can also take a hybrid approach
\cite{hwang2017character,kawakami-dyer-blunsom:2017:Long},
recent advances indicate that full
character-level modeling is now competitive with word-level modeling.
A large
part of this is due to the use of recurrent neural networks
\cite{DBLP:conf/interspeech/MikolovKBCK10}, which
can generalize about how the distribution $p(c_i \mid \charseq_{<i})$ depends on $\charseq_{<i}$.

We use a long short-term memory (LSTM) \LM \cite{sundermeyer2012lstm}, identical to that of \newcite{,DBLP:journals/corr/ZarembaSV14}, but at the character-level.
To achieve the hidden state $\mathbf{h}_i \in \mathbb{R}^d$ at time step $i$, one
feeds the left context $\charseq_{i-1}$ to the LSTM:
 $\mathbf{h}_i = \text{LSTM}\left( c_1, \ldots, c_{i-1} \right)$
where the model uses a learned vector to represent each character type.
This involves a recursive procedure described in
\newcite{hochreiter1997long}. Then, the probability
distribution over the $i^\text{th}$ character is
$p(c_t \mid \charseq_{< i}) = \text{softmax}\left(\mathbf{W}\, \mathbf{h}_i + \mathbf{b}\right)$,
where $\mathbf{W} \in \mathbb{R}^{|\Sigma|\times d}$ and $\mathbf{b} \in \mathbb{R}^{|\Sigma|}$ are parameters.

Parameters for all models are estimated on the training
portion and model selection is performed on the development portion. The
neural models are trained with SGD \cite{robbins1951stochastic} with
gradient clipping, such that each component has a maximum absolute
value of $5$. We optimize for 100 iterations and perform early stopping (on the development portion).
We employ a character embedding of size $1024$ and 2
hidden layers of size $1024$.\footnote{As \newcite{DBLP:journals/corr/ZarembaSV14} indicate, increasing the number of parameters may allow us to achieve better
performance.}  The implementation is in
PyTorch.

\section{A Fairer Evaluation: Multi-Text}\label{sec:evaluation}
Effecting a cross-linguistic study on \LMs is complicated
because different models could be trained and tested on incomparable
corpora.  To avoid this problem, we use \textbf{multi-text}: $k$-way translations of the same semantic content.

\paragraph{What's wrong with bits per character?}
Open-vocabulary language modeling is most commonly evaluated under
\textbf{bits per character} (BPC) $ = \frac{1}{|\charseq|+1} \sum^{|\charseq|+1}_{i=1} \log p(c_i \mid \charseq_{< i})$.%
\footnote{To aggregate this over an entire test corpus, we replace the
denominator and also the numerator by summations over all
utterances $\charseq$.}
Even with multi-text,
comparing BPC is not straightforward, as it relies on the vagaries of
individual writing systems. Consider, for example, the
difference in how Czech and German express the phoneme /\word{t\textipa{S}}/:
Czech uses \word{\v{c}}, whereas German \word{tsch}. Now, consider the
Czech word \word{pu\v{c}} and its German equivalent
\word{Putsch}. Even if these words are both predicted with the
\emph{same} probability in a given context, German will end up with a
lower BPC.
\footnote{Why not work with \emph{phonological} characters, rather
than orthographic ones, obtaining /\word{put\textipa{S}}/ for both Czech and German?
Sadly this option is also fraught with problems
as many languages have perfectly predictable phonological elements that will
artificially lower the score.}

\paragraph{Bits per English Character.}
Multi-text allows us to compute a fair metric
that is invariant to the orthographic (or phonological)
changes discussed above: \textbf{bits per English character} (BPEC).
  $\text{BPEC} = \frac{1}{|\charseq_{\textit{English}}|+1} \sum^{|\charseq|+1}_{i=1} \log p(c_i \mid \charseq_{< i})$,
where $\charseq_\textit{English}$ is
the English character sequence in the utterance aligned to
$\charseq$. The choice of English is arbitrary, as any
other choice of language would simply scale the values by a constant factor.

Note that this metric is essentially capturing the overall \emph{bits per utterance}, and that normalizing using English characters only makes numbers independent of the overall utterance length; it is not critical to the analysis we perform in this paper.

\paragraph{A Potential Confound: Translationese.}
Working with multi-text, however, does introduce a new bias: all of the utterances in the corpus have a source language and 20 translations of that source utterance into target languages.  The characteristics of translated language has been widely studied and exploited, with one prominent characteristic of translations being simplification \cite{baker1993corpus}.

Note that a significant fraction of the
original utterances in the corpus are English.  Our analysis may then
have underestimated the BPEC for other languages, to the extent that
their sentences consist of simplified ``translationese.''  Even so,
English had the lowest BPEC from among the set of languages.

\begin{table}
  \begin{adjustbox}{width=\columnwidth}
    \begin{tabular}{@{~}c@{~}|@{~}c@{~}|@{~}c@{~}|@{~}c@{~}|@{~}c@{~}|@{~}c@{~}|@{~}c@{~}} \toprule
          \multicolumn{3}{@{~}c@{~}|}{~} & \multicolumn{4}{c}{BPEC~/~$\Delta$BPC ($\cdot$e-2)}   \\ \midrule
          & data (M) &  & \multicolumn{2}{c|}{hybrid $n$-gram}  & \multicolumn{2}{c}{LSTM} \\ \midrule
      lang & wds~/~ch  & MCC & form  & lemma  & form  & lemma \\ \midrule
      bg & 0.71/4.3 & 96  & 1.13/\textcolor{darkblue}{\phantom{1}\phantom{-}4} & 1.03/\textcolor{darkblue}{\phantom{1}\phantom{-}1} & 0.95/\textcolor{darkblue}{\phantom{1}\phantom{-}3} & 0.80/\textcolor{darkblue}{\phantom{1}\phantom{-}1} \\
  cs & 0.65/3.9 & 195 & 1.20/\textcolor{darkred}{\phantom{1}-8} & 1.05/\textcolor{darkred}{-12} & 0.97/\textcolor{darkred}{\phantom{1}-6} & 0.83/\textcolor{darkred}{\phantom{1}-9} \\
  da & 0.70/4.1 & 15 & 1.10/\textcolor{darkred}{\phantom{1}-1} & 1.06/\textcolor{darkred}{\phantom{1}-4} & 0.85/\textcolor{darkred}{\phantom{1}-1} & 0.82/\textcolor{darkred}{\phantom{1}-3} \\
  de & 0.74/4.8 & 38 & 1.25/\textcolor{darkblue}{\phantom{-}17} & 1.18/\textcolor{darkblue}{\phantom{-}13} &  1.04/\textcolor{darkblue}{\phantom{-}14} & 0.90/\textcolor{darkblue}{\phantom{-}10} \\
  el & 0.75/4.6 & 50 & 1.18/\textcolor{darkblue}{\phantom{-}13} & 1.08/\textcolor{darkblue}{\phantom{1}\phantom{-}5} & 0.90/\textcolor{darkblue}{\phantom{-}10} & 0.82/\textcolor{darkblue}{\phantom{1}\phantom{-}4} \\
  en & 0.75/4.1 & 6 & 1.10/\phantom{1}\phantom{-}0 & 1.08/\textcolor{darkred}{\phantom{1}-3} & 0.85/\phantom{1}\phantom{-}0 & 0.83/\textcolor{darkred}{\phantom{1}-3} \\
  es & 0.81/4.6 & 71 & 1.15/\textcolor{darkblue}{\phantom{-}12} & 1.07/\textcolor{darkblue}{\phantom{1}\phantom{-}7} & 0.87/\textcolor{darkblue}{\phantom{1}\phantom{-}9} & 0.80/\textcolor{darkblue}{\phantom{1}\phantom{-}5} \\
  et$^*$ & 0.55/3.9 & 110 & 1.20/\textcolor{darkred}{\phantom{1}-8} & 1.11/\textcolor{darkred}{-15} & 0.97/\textcolor{darkred}{\phantom{1}-6} & 0.89/\textcolor{darkred}{-12} \\
  fi$^*$ & 0.52/4.2 & 198 & 1.18/\textcolor{darkblue}{\phantom{1}\phantom{-}2} & 1.02/\textcolor{darkred}{-11} & 1.05/\textcolor{darkblue}{\phantom{1}\phantom{-}1} & 0.79/\textcolor{darkred}{\phantom{1}-9} \\
  fr & 0.88/4.9 & 30 & 1.13/\textcolor{darkblue}{\phantom{-}17} & 1.06/\textcolor{darkblue}{\phantom{-}13} & 0.92/\textcolor{darkblue}{\phantom{-}14} & 0.78/\textcolor{darkblue}{\phantom{-}10} \\
  hu$^*$ & 0.63/4.3 & 94 & 1.25/\textcolor{darkblue}{\phantom{1}\phantom{-}5} & 1.12/\textcolor{darkred}{\phantom{1}-9} & 1.09/\textcolor{darkblue}{\phantom{1}\phantom{-}5} & 0.89/\textcolor{darkred}{\phantom{1}-7} \\
  it & 0.85/4.8 & 52 & 1.15/\textcolor{darkblue}{\phantom{-}16} & 1.08/\textcolor{darkblue}{\phantom{-}14} & 0.96/\textcolor{darkblue}{\phantom{-}14} & 0.79/\textcolor{darkblue}{\phantom{-}10} \\
  lt & 0.59/3.9 & 152 & 1.17/\textcolor{darkred}{\phantom{1}-6} & 1.12/\textcolor{darkred}{\phantom{1}-7} & 0.93/\textcolor{darkred}{\phantom{1}-5} & 0.88/\textcolor{darkred}{\phantom{1}-6} \\
  lv & 0.61/3.9 & 81 & 1.15/\textcolor{darkred}{\phantom{1}-6} & 1.04/\textcolor{darkred}{\phantom{1}-9} & 0.91/\textcolor{darkred}{\phantom{1}-5} & 0.81/\textcolor{darkred}{\phantom{1}-7} \\
  nl & 0.75/4.5 & 26 & 1.20/\textcolor{darkblue}{\phantom{-}11} & 1.16/\textcolor{darkblue}{\phantom{1}\phantom{-}4} & 0.92/\textcolor{darkblue}{\phantom{1}\phantom{-}8} & 0.91/\textcolor{darkblue}{\phantom{1}\phantom{-}4} \\
  pl & 0.65/4.3 & 112 & 1.21/\textcolor{darkblue}{\phantom{1}\phantom{-}6} & 1.09/\textcolor{darkred}{\phantom{1}-1} & 0.97/\textcolor{darkblue}{\phantom{1}\phantom{-}5} & 0.84/\textcolor{darkred}{\phantom{1}-1} \\
  pt & 0.89/4.8 & 77 & 1.17/\textcolor{darkblue}{\phantom{-}16} & 1.09/\textcolor{darkblue}{\phantom{1}\phantom{-}9} & 0.88/\textcolor{darkblue}{\phantom{-}12} & 0.82/\textcolor{darkblue}{\phantom{1}\phantom{-}7} \\
  ro & 0.74/4.4 & 60 & 1.17/\textcolor{darkblue}{\phantom{1}\phantom{-}8} & 1.09/\phantom{1}\phantom{-}0 & 0.90/\textcolor{darkblue}{\phantom{1}\phantom{-}6} & 0.84/\phantom{1}\phantom{-}0 \\
  sk & 0.64/3.9 & 40 & 1.16/\textcolor{darkred}{\phantom{1}-6} & 1.06/\textcolor{darkred}{-11} & 0.92/\textcolor{darkred}{\phantom{1}-5} & 0.87/\textcolor{darkred}{\phantom{1}-9} \\
  sl & 0.64/3.8 & 100 & 1.15/\textcolor{darkred}{-10} & 1.02/\textcolor{darkred}{-10} & 0.90/\textcolor{darkred}{\phantom{1}-8} & 0.80/\textcolor{darkred}{\phantom{1}-7} \\
  sv & 0.66/4.1 & 35 & 1.11/\textcolor{darkred}{\phantom{1}-2} & 1.06/\textcolor{darkred}{\phantom{1}-8} & 0.86/\textcolor{darkred}{\phantom{1}-2} & 0.83/\textcolor{darkred}{\phantom{1}-7} \\
   \bottomrule
    \end{tabular}
    \end{adjustbox}
    \caption{\footnotesize Results for all configurations and the typological profile of the 21 Europarl languages.  All languages are Indo-European, except for those marked with $^*$ which are Uralic.  Morpholical counting complexity (MCC) is given for each language, along with bits per English character (BPEC) and the $\Delta$BPC, which is BPEC minus bits per character (BPC).  This is \textcolor{darkblue}{blue} if BPEC $>$ BPC and \textcolor{darkred}{red} if BPEC $<$ BPC.}
  \label{tab:bigv2}
\end{table}

\section{Experiments and Results}\label{sec:experiments}
  Our experiments are conducted on the 21 languages of the Europarl
  corpus \cite{koehn2005europarl}.  The corpus consists of
  utterances made in the European parliament and are
  aligned cross-linguistically by a unique utterance id.  With the exceptions (noted in \cref{tab:bigv2}) of Finnish, Hungarian
and Estonian, which are Uralic, the languages are Indo-European.

While Europarl does not contain quite our desired breadth of typological diversity, it serves our purpose by providing large collections of aligned data across many languages.
 To create
  our experimental data, we extract all utterances and randomly sort
  them into train-development-test splits such that roughly 80\% of
  the data are in train and 10\% in development and test,
  respectively.\footnote{Characters appearing $< 100$ times in train are $\question$.} We also perform experiments
  on {\em lemmatized} text, where we replace every word with its lemma
  using the UDPipe toolkit \cite{straka2016udpipe}, stripping away its
  inflectional morphology.  We report two evaluation metrics: BPC
  and BPEC (see \cref{sec:evaluation}).
  Our BPEC measure always normalizes by the length of the original, not lemmatized, 
  English.

Experimentally, we want to show: (i) When evaluating models in a controlled environment (multi-text under BPEC), the models achieve lower performance on certain languages and (ii)
inflectional morphology is the primary culprit for the performance
differences. 
 However, we repeat that we do not in this paper tease apart whether
the models are at fault, or that certain languages inherently encode more information.

\section{Discussion and Analysis}
We display the performance of the $n$-gram \LM and the LSTM \LM under BPC
and BPEC for each of the 21 languages in \cref{fig:graphs} with full
numbers listed in \cref{tab:bigv2}. There are several main take-aways.

\paragraph{The Effect of BPEC.}
The first major take-away is that BPEC offers a
cleaner cross-linguistic comparison than BPC.
Were we to rank the languages
by BPC (lowest to highest), we would find
that English was in the middle of the pack, which is surprising
as new language models are often only tuned on English itself.
For example, BPC surprisingly suggests that French is easier to model
than English.  However, ranking under BPEC shows that the LSTM has the easiest
time modeling English itself.
Scandinavian
languages Danish and Swedish have BPEC closest to
English; these languages are typologically and genetically similar to English.

\paragraph{$n$-gram versus LSTM.}  As expected, the LSTM outperforms
the baseline $n$-gram models across the board. In addition, however,
$n$-gram modeling yields relatively poor performance on some
languages, such as Dutch, with only modestly more complex inflectional
morphology than English.  Other phenomena---e.g., perhaps,
compounding---may also be poorly modeled by $n$-grams.

\begin{figure}
   \centering
   \includegraphics[width=\columnwidth]{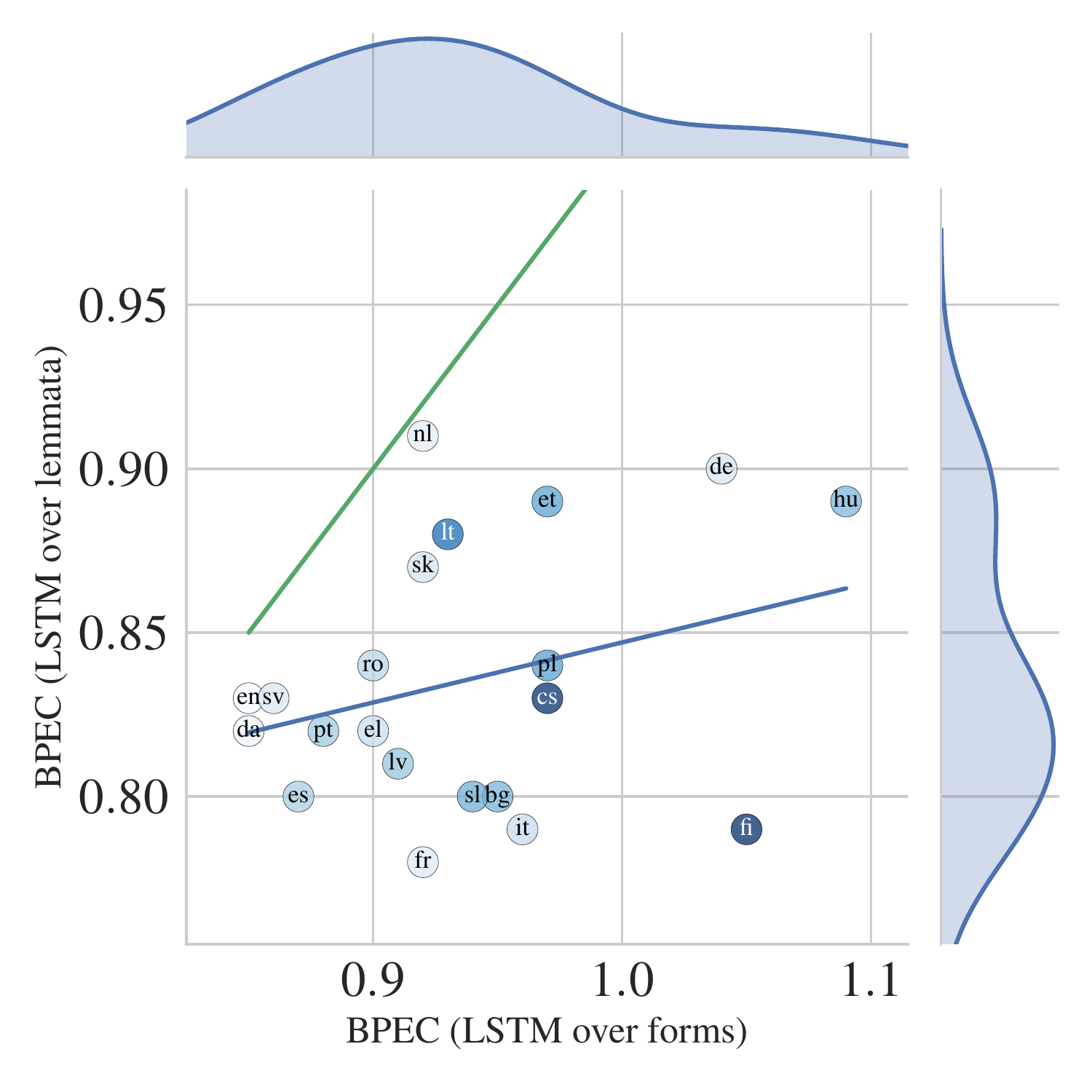}\vspace*{-0.15in}
   \caption{Each dot is a language, and its coordinates are the BPEC values
     for the LSTM \LM{}s over words and lemmata.  The top and right margins
     show kernel density estimates of these two sets of BPEC values.
     All dots follow the blue regression, but stay
     below the green line ($y=x$), and the darker dots---which
     represent languages with higher counting complexity---tend to fall toward the right but
     not toward the top, since counting complexity is correlated only
     with the BPEC over words.}
   \label{fig:lemmata-v-forms}
 \end{figure}

\paragraph{The Impact of Inflectional Morphology.}
Another major take-away is that rich inflectional morphology is a
difficulty for both $n$-gram and LSTM LMs.  In this section we
give numbers for the LSTMs.  Studying
\cref{fig:words}, we find that Spearman's rank correlation between a
language's BPEC and its counting complexity
(\cref{sec:inflection}) is quite high ($\rho=0.59$, significant at
$p < 0.005$).  This clear correlation between the level of
inflectional morphology and the LSTM performance indicates that
character-level models do not automatically fix the problem of
morphological richness. If we lemmatize the words, however
(\cref{fig:lemmata}), the correlation becomes insignificant and in
fact slightly negative ($\rho=-0.13$, $p \approx 0.56 $).
The difference of the two previous graphs (\cref{fig:diff}) shows
more clearly that the LM penalty for modeling inflectional endings is
greater for languages with higher counting complexity. Indeed, this penalty is arguably a more appropriate measure of the complexity of the
inflectional system.  See also \cref{fig:lemmata-v-forms}.

The differences in BPEC among languages are reduced when we lemmatize, with standard deviation dropping from $0.065$ bits to $0.039$ bits.  Zooming in on Finnish (see \cref{tab:bigv2}), we see that Finnish forms are harder to model than English forms, but Finnish lemmata are \emph{easier} to model than English ones. This is strong evidence that it was primarily the inflectional morphology, which lemmatization strips, that caused the differences in the model's performance on these two languages.

\section{Related Work}

Recurrent neural language models can effectively learn
complex dependencies, even in open-vocabulary settings
\cite{hwang2017character,kawakami-dyer-blunsom:2017:Long}.  Whether
the models are able to learn particular syntactic interactions is an
intriguing question, and some methodologies have been presented to
tease apart under what circumstances variously-trained models encode
attested interactions
\cite{LinzenEtAl-2016:TACL,EnguehardEtAl-2017:CoNLL}. While the sort
of detailed, construction-specific analyses in these papers is surely
informative, our evaluation is language-wide.

MT researchers have investigated whether an
English sentence contains enough information to predict the
fine-grained inflections used in its foreign-language translations
\citep[see][]{kirov-et-al-2017}.

\newcite{sproat2014database} present a corpus of close translations of
sentences in typologically diverse languages along with detailed
morphosyntactic and morphosemantic annotations, as the means for
assessing linguistic complexity for comparable messages, though they
expressly do not take an information-theoretic approach to measuring
complexity. In the linguistics literature,
\newcite{mcwhorter2001world} argues that certain languages are less
complex than others: he claims that Creoles are
simpler. \newcite{mueller-schuetze-schmid:2012:NAACL-HLT} compare \LMs
on EuroParl, but do not compare performance across languages.

\vspace{-1pt}
\section{Conclusion}
\vspace{-3pt}

We have presented a clean method for the cross-linguistic comparison
of language modeling: We assess whether a language modeling technique
can compress a sentence and its translations equally well.
We show an interesting correlation between the morphological
richness of a language and the performance of the model. In an attempt
to explain causation, we also run our models on lemmatized versions of
the corpora, showing that, upon the removal of inflection, no such
correlation between morphological richness and \LM performance exists.
It is still unclear, however, whether the performance difference
originates from the inherent difficulty of the languages or with the
models.

\bibliography{complexity}
\bibliographystyle{acl_natbib}

\end{document}